\documentclass{bmcart}

\RequirePackage{natbib}
\usepackage[utf8]{inputenc} 
\usepackage{graphicx}

\startlocaldefs
\endlocaldefs

\begin{document}

\begin{frontmatter}

\begin{fmbox}
\dochead{Research}


\title{A hybrid quantum-classical conditional generative adversarial network algorithm for human-centered paradigm in cloud}


\author[
   addressref={aff1},                   
   email={wenjiel@163.com}   
]{\inits{WJL}\fnm{Wenjie} \snm{Liu}}
\author[
   addressref={aff2},
   corref={aff1},                       
   email={1175666021@qq.com}
]{\inits{YZ}\fnm{Ying} \snm{Zhang}}
\author[
   addressref={aff2},
   email={mtsdzl@163.com}
]{\inits{ZLD}\fnm{Zhiliang} \snm{Deng}}
\author[
   addressref={aff1},
   email={2759312576@qq.com}
]{\inits{JJZ}\fnm{Jiaojiao} \snm{Zhao}}
\author[
   addressref={aff3}
]{\inits{LT}\fnm{Lian} \snm{Tong}}

\address[id=aff1]{
  \orgname{Nanjing University of Information Science \& Technology, School of Computer and Software}, 
  \street{Ningliu Road},                     %
  \postcode{210044}                                
  \city{Nanjing},                              
  \cny{China}                                    
}
\address[id=aff2]{%
  \orgname{Nanjing University of Information Science \& Technology, School of Automation},
  \street{Ningliu Road},
  \postcode{210044}
  \city{Nanjing},
  \cny{China}
}
\address[id=aff3]{%
  \orgname{School of Information Engineering, Jiangsu Maritime Institute},
  \postcode{211100}
  \city{Nanjing},
  \cny{China}
}


\end{fmbox}


\begin{abstractbox}

\begin{abstract} 
 As an emerging field that aims to bridge the gap between human activities and computing systems, human-centered computing (HCC) in cloud, edge, fog has had a huge impact on the artificial intelligence algorithms. The quantum generative adversarial network (QGAN) is considered to be one of the quantum machine learning algorithms with great application prospects, which also should be improved to conform to the human-centered paradigm. The generation process of QGAN is relatively random and the generated model does not conform to the human-centered concept, so it is not quite suitable for real scenarios. In order to solve these problems, a hybrid quantum-classical conditional generative adversarial network (QCGAN) algorithm is proposed, which is a knowledge-driven human-computer interaction computing mode in cloud. The purpose of stabilizing the generation process and realizing the interaction between human and computing process is achieved by inputting artificial conditional information in the generator and discriminator. The generator uses the parameterized quantum circuit with an all-to-all connected topology, which facilitates the tuning of network parameters during the training process. The discriminator uses the classical neural network, which effectively avoids the "input bottleneck" of quantum machine learning. Finally, the BAS training set is selected to conduct experiment on the quantum cloud computing platform. The result shows that the QCGAN algorithm can effectively converge to the Nash equilibrium point after training and perform human-centered classification generation tasks.

\end{abstract}


\begin{keyword}
\kwd{Quantum generative adversarial network}
\kwd{Conditional generative adversarial network}
\kwd{Human-centered computing}
\kwd{Cloud computing}
\kwd{Parameterized quantum circuits}
\end{keyword}


\end{abstractbox}
%

\end{frontmatter}



\section{Introduction}

With the development of wireless communications and networking, human-centered computing (HCC) in cloud, edge, and fog attempts to effectively integrate various computing elements related to humans \cite{Singh,ZhouC}, which becomes a common focus of attention in the academic and industrial fields. Unlike other ordinary computing, HCC pays more attention to the status of human in computing technology and the interaction of humans with cyberspace and physical world \cite{LiuT}. Therefore, the design of HCC systems and algorithms needs to take into account the individual's ability and subjective initiative \cite{QiL,Srivastava}.
Among them, cloud computing uses a super-large-scale distributed computing method to adapt to the large number of examples and complex calculation requirements of current artificial intelligence (AI) algorithms, and it has become a computing method commonly sought \cite{XuX,Khan}.
In the background of HCC computing and big data, there are many interesting and practical applications generating \cite{WangL,XuY,LiJ}. Privacy is also an important norm that computing models must pay attention to, especially related to privacy perception and privacy protection \cite{QiLY,ZhongW,LiuQ}.

Quantum cloud computing allows users to test and develop their quantum programs on local personal computers, and run them on actual quantum devices, thereby reducing the distance between humans and the mysterious quantum \cite{Dumitrescu}. Under the influence of the AI wave, many technology companies are committed to establishing quantum cloud computing platforms that enable users to implement quantum machine learning algorithms. Compared with the two major models of machine learning, the generative model and the discriminant model, the generative model is more capable of exerting human subjective initiative, so it has the potential to developed into the HCC paradigm. Therefore, we consider the very creative quantum generative adversarial network model as a breakthrough in HCC computing in cloud .

Generative adversarial network (GAN) \cite{Good} evaluates generative models through a set of adversarial neural network frameworks, which is a hot topic in recent years about generative machine learning algorithm. The GAN algorithm is bases on game theory scenario, and the generator aims to learn the mapping from simple input distribution to complex training sample space by competing with discriminator. As the adversary, the discriminator should judge as accurately as possible whether the input data comes from the training set or the generator. Both participants of the game try to minimize their own loss, so that the adversarial network framework finally reaches the Nash equilibrium \cite{WangK}. In recent years, GAN has been successfully used in the fields of the processing of image, audio, natural language etc., to achieve functions such as clear image generation \cite{Ledig,Kupyn}, video prediction \cite{ZhuL}, text summarization  \cite{Bhargava}, and image generation of semantic \cite{Johnson}. Actually, it is difficult to ensure stable training of GAN in operation. Researchers use the relevant results obtained by deep learning to improve GAN, including methods such as designing new network structures \cite{Radford}, adding regular constraints \cite{Ioffe}, integrated learning \cite{Tolstikhin}, and improving optimization algorithms \cite{1701.07875}. However, the improved algorithms above are not human-centered, because the rules learned by the GAN algorithm are implicit. It is difficult to generate data that meets specific requirements by changing the structure or input of a trained generator. In 2014, Mirza et al. proposed conditional generative adversarial network (CGAN) \cite{Mirza}. This method guides GAN to learn to sample from the conditional distribution by adding conditional constraints to the hidden variables of the input layer, so that the generative data can be guided by conditional inputs, thereby expanding the application scenarios of the GAN algorithm. In the construction, the setting of conditional constraints can make the subjective initiative of people play a role, so it can be regarded as an HCC algorithm. Based on the CGAN algorithm, many human-centered applications have been constructed, such as objects detection \cite{ZhuD}, medical images processing and synthesis \cite{Dar,YiX}.

Quantum generative adversarial network (QGAN) is a data-driven quantum circuit machine learning algorithm which combine the classical GAN and quantum computing \cite{Marce}. In 2018, Lloyd proposed the concept of QGAN \cite{Seth}, which analyzed the effectiveness of three different QGAN frameworks from a theoretical perspective, and demonstrated that quantum adversarial learning can also reach the Nash equilibrium point when the generative distribution can fit real distribution. In the same year, Pierre's team discussed QGAN in more detail, by giving the general structure of the parameterized quantum circuit (PQC) as a generator and the estimation method of the parameter gradient when training the network \cite{Dalla}. In 2019, Hu et al. used quantum superconducting circuit physics experiments to prove the feasibility of QGAN on current noisy intermediate-scale quantum (NISQ) devices \cite{HuL}. Additionally, the optimization of the quantum generator structure is also one of the research priorities. For example, using matrix product state \cite{HanZ} and tree tensor network \cite{Hugg} to construct PQCs as generator and discriminator of GAN respectively, the convergence and robustness to noise of these methods are all verified through experiments on quantum hardware.

In terms of generating quantum data, the quantum supremacy means that classical information processors or neural networks sometimes cannot fit the data generated by quantum systems, and only quantum generator can complete such tasks. For the generation of classical data, the output of quantum generator can always meet the differentiable constraint. By sampling the output of quantum generator, classical discrete data can be obtained. In contrast, classical GAN cannot directly generate discrete data due to the influence of differentiable constraint. Therefore, as a complement to the classical GAN, QGAN with the ability to generate discrete data and the combination of other known variants of GAN and quantum mechanical mechanisms are of great research value.

Similar to classical GAN, QGAN also has the problem of uncontrollable training process and random generative output. However, in practical applications, the intent output obtained by changing the input is a more common situation, so QGAN is less practical. In order to solve the problem that the QGAN algorithm lacks human-oriented thinking, this paper proposes a hybrid quantum classical scheme based on conditional generation adversarial network. Conditional constraints are added to the QGAN algorithm to guide the training process. This method has both the controllability of CGAN and the discrete data generation capability of QGAN. By analyzing the performance of different GAN, it is proved that the algorithm is better than the classical CGAN in terms of time complexity and algorithm functions. Through modeling and training experiments in cloud on classical data generation problem, the convergence of the model and the accuracy of the generative data verify the feasibility of applying quantum computing to the CGAN structure.

The rest of the paper is organized as follows. Section II describes the preliminaries about classical GAN and QGAN. Section III presents the design focus of QCGAN, including the method of designing the PQCs and estimating the parameter gradients. The performance analysis of QCGAN and the comparison with other related algorithms are in Section IV. In Section V, experiments are performed in the quantum cloud computing platform to verify the feasibility of the proposed QCGAN algorithm. Section VI summarizes what we find in this work and the prospects for future researches.

\section{Principles of generative adversarial network algorithm}

\subsection{Generative adversarial network}
The core idea of the classical GAN is to construct a zero-sum game between generator and discriminator. Through the adversarial learning strategy, generator and discriminator are alternately trained to obtain a better generative model. The structure and algorithm flowchart of GAN are shown in Fig. 1.

  \begin{figure}[h!]
  \includegraphics[width=4.5in]{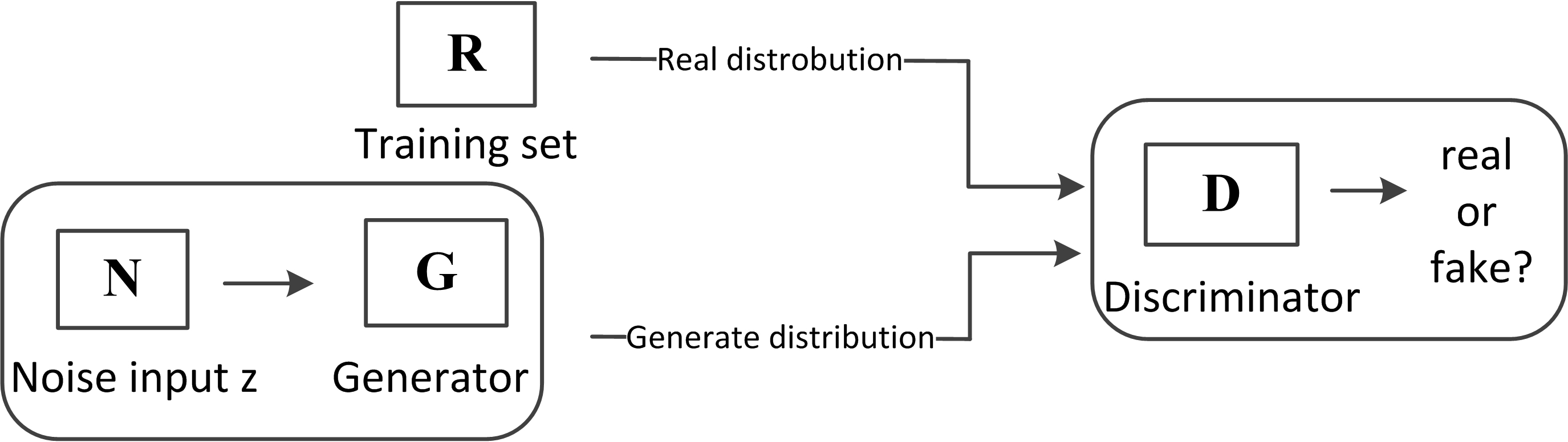}
  \caption{\csentence{Schematic diagram of classical generative adversarial network.}}
      \end{figure}

Specifically, the first step is to give training samples as generation target, assuming that the real data comes from a fixed and unknown distribution ${p_{real}}\left( x \right)$. The generator is a neural network that can map low-dimensional distribution to high-dimensional space, and the discriminator is a neural network with classification function. The parameters of generator and discriminator are denoted as ${\overrightarrow \theta  _G}$ and ${\overrightarrow \theta  _D}$, respectively. The input of generator is a noise vector $z$, which is generally sampled from a normal distribution or a uniform distribution; $x = G\left( {{{\overrightarrow \theta  }_G},z} \right)$ is the output of generator, which is transformed from the noise vector, and constitutes the generative distribution ${p_G}\left( x \right)$. In the case of completing the ideal adversarial training, the discriminator will not be able to distinguish whether the input comes from the real distribution ${p_{real}}\left( x \right)$ or the generative distribution ${p_G}\left( x \right)$. Therefore, the goal of training generator is to make discriminator distinguish the output of generator as real data as much as possible. On the other hand, when training discriminator, its input contains real data $x \sim {p_{real}}\left( x \right)$ and the output of generator $x \sim {p_G}\left( x \right)$. At this time, the training goal is to accurately judge the two categories of input data. Combining these two aspects, the optimization of GAN can be described as the following minimax game problem
\begin{eqnarray}\label{eqexpmuts}
\mathop {\min }\limits_G \mathop {\max }\limits_D V\left( {D,G} \right) = {E_{x \sim {p_{real}}}}\left[ {\log D\left( x \right)} \right] + E{}_{x \sim {p_G}}\left[ {\log \left( {1 - D\left( x \right)} \right)} \right].
\end{eqnarray}

\subsection{Conditional generative adversarial network}
In view of the uncontrollable shortcoming of the training process of GAN, the CGAN algorithm adds conditional variables to the input of generator and discriminator at the same time to play a constraining and guiding role. The structure and flowchart of CGAN algorithm are shown in Fig. 2. The condition variables $y$ are generally known information with specific semantics, such as feature labels. Under the CGAN framework, the generator pays more attention to sample features that are closely related to conditional constraints, ignores other less relevant local features. Therefore, the addition of condition variables can control the training process to generate higher quality data. The output of the generator can be regarded as sampling from the conditional distribution ${p_G}\left( {x\left| y \right.} \right)$, so the objective function of CGAN can be rewritten on the basis of the original GAN as
\begin{eqnarray}\label{eqexpmuts}
\mathop {\min }\limits_G \mathop {\max }\limits_D V\left( {D,G} \right) = {E_{x \sim {p_{real}}}}\left[ {\log D\left( {x\left| y \right.} \right)} \right] + E{}_{x \sim {p_G}}\left[ {\log \left( {1 - D\left( {x\left| y \right.} \right)} \right)} \right].
\end{eqnarray}

  \begin{figure}[h!]
  \includegraphics[width=4.5in]{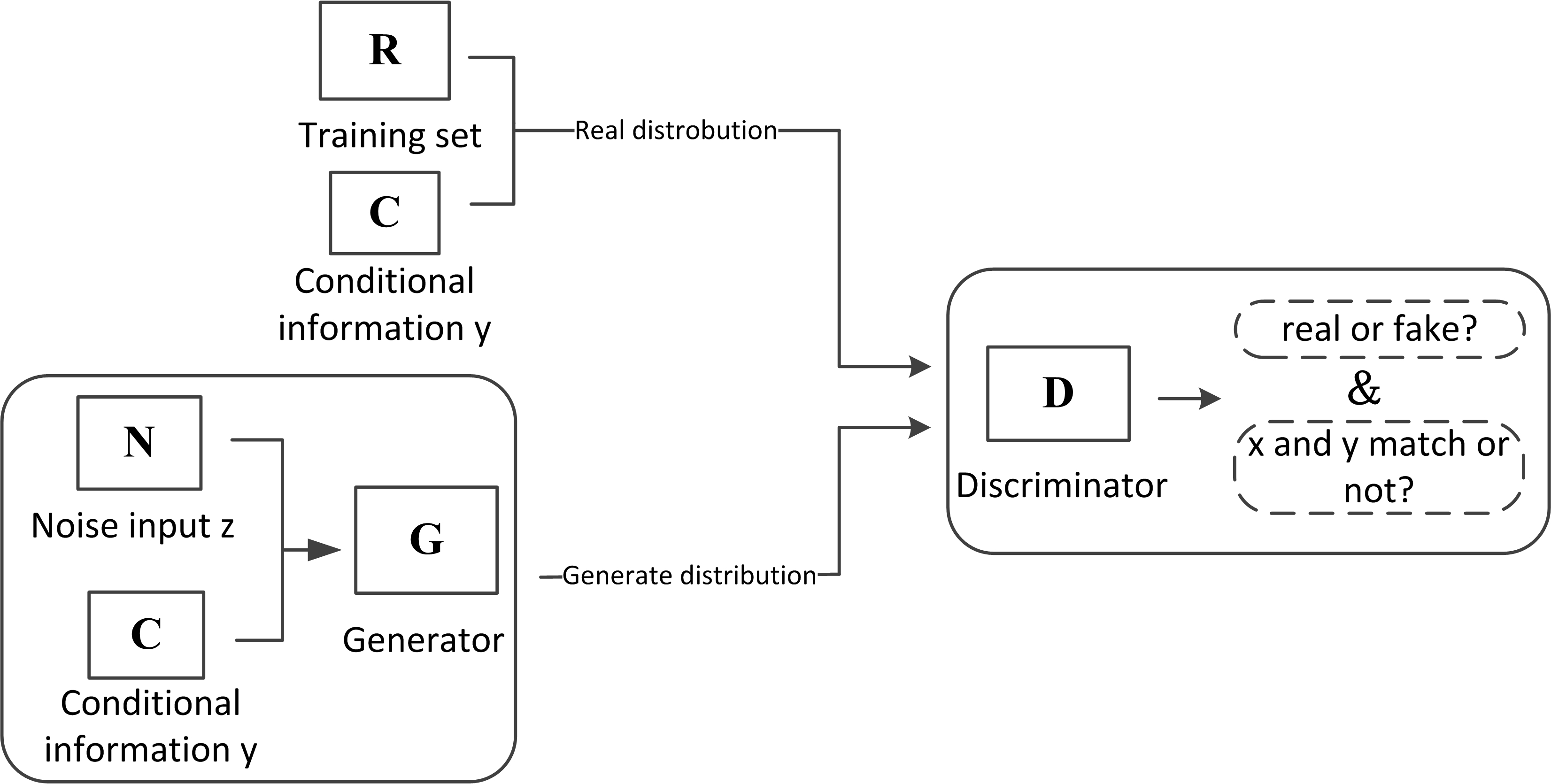}
  \caption{\csentence{Schematic diagram of classical conditional generative adversarial network.}}
      \end{figure}

CGAN needs to sample from the noise vector and the condition variable at the same time, so the set of reasonable condition variable according to the generation target plays a crucial role in the generator's ability to fit the real distribution. The most common method is to directly extract the conditional variables from the training data, so that generator and discriminator get some prior knowledge about the training set when they receive the input. For example, the category label is used as a conditional variable and attached to the input layer of the adversarial network \cite{Mirza}. At this time, CGAN can be regarded as an improvement of the unsupervised GAN model into a weakly supervised or a supervised model.

\subsection{Quantum generative adversarial network}
The QGAN is also a zero-sum game that constructed by generator and discriminator in principle. If one or more than one of the real data, the generator and the discriminator obey the quantum mechanism, the constructed algorithm scheme belongs to the QGAN concept. In general, the quantum data set is expressed in the form of a density matrix, which corresponds to the covariance matrix of the classical data set. Quantum generator and discriminator are composed of PQC. The selection, arrangement, and depth of quantum gates of PQC will affect the performance of it, so they are also the parts that can be optimized.

When QGAN is used for classical data generation tasks, if the goal of the generator is to reproduce statistical data on high-dimensional, QGAN with quantum generator has the potential to exponentially accelerate the convergence to Nash equilibrium \cite{Seth}. Using classical neural networks as the discriminator in adversarial learning can avoid the input bottleneck of quantum machine learning, because it reduces the calculation and resource consumption of quantum state encoding when discriminate real classical data. Combining the above two aspects, the QCGAN algorithm proposed in this paper is based on the basic settings of the quantum generator and the classical discriminator to generate classical data. The structure and algorithm flowchart of this kind of QGAN algorithm are shown in Fig. 3.

  \begin{figure}[h!]
  \includegraphics[width=4.5in]{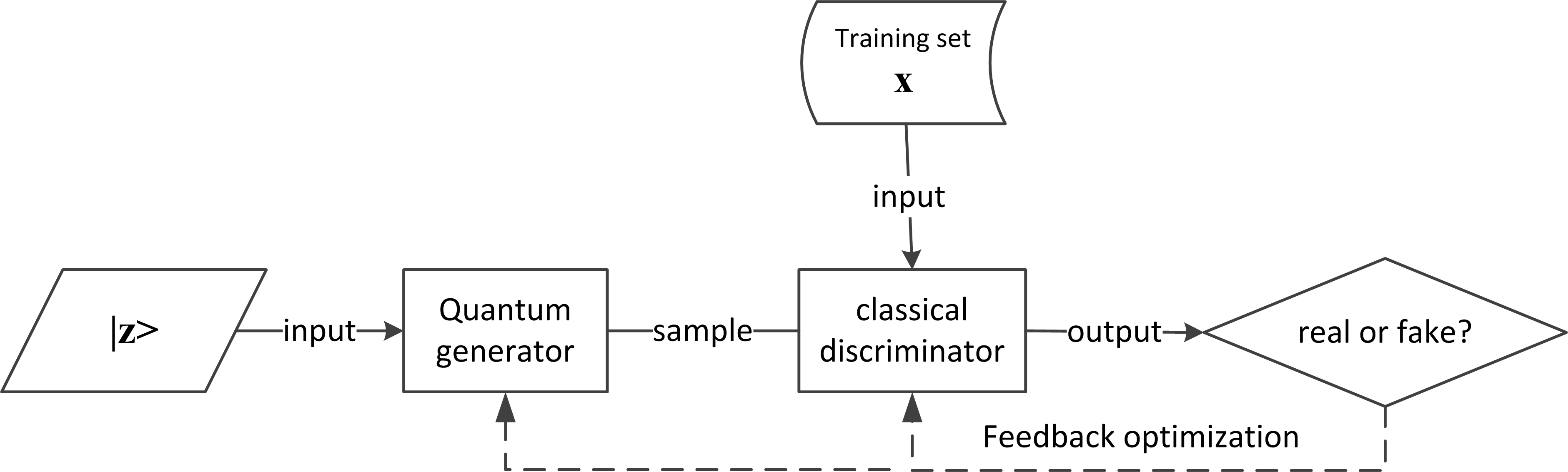}
  \caption{\csentence{Schematic diagram of quantum generative adversarial network.}}
      \end{figure}

\section{Quantum conditional generative adversarial network algorithm}
The QCGAN algorithm proposed in this paper is a generative adversarial network model which is suitable for fitting classical data distribution, whose generation process is controllable. The generator of QCGAN is constructed in the form of the parameterized quantum circuit, and the discriminator uses a classical neural network to complete the classification task. Different from the unconstrained QGAN algorithm, the QCGAN algorithm adds conditional variables to the input of both generator and discriminator to guide the training process. The basic flow of the algorithm can be summarized as follows (as shown in Fig. 4): the first step is to prepare classical samples and introduce appropriate conditional constraints according to the data characteristics as well as the goal of generation task. This two parts are combined to form the training data set of the network. The classical conditional constraints, which reflect the statistical characteristics of the training data set, are encoded into a entangled quantum state through a well-designed quantum circuit. The next step is to construct the PQC of the generator and the classical neural network of discriminator. Finally, the generative distribution and the real distribution are sampled separately and input these data to the discriminator for classification, and then an adversarial strategy is formulated for training. If the objective function converges, it means finding the best quantum generator. The output of the generator can be sampled to get a set of classical data, which is the result not only fits the target distribution but also meets the constraints.

  \begin{figure}[h!]
  \includegraphics[width=4.5in]{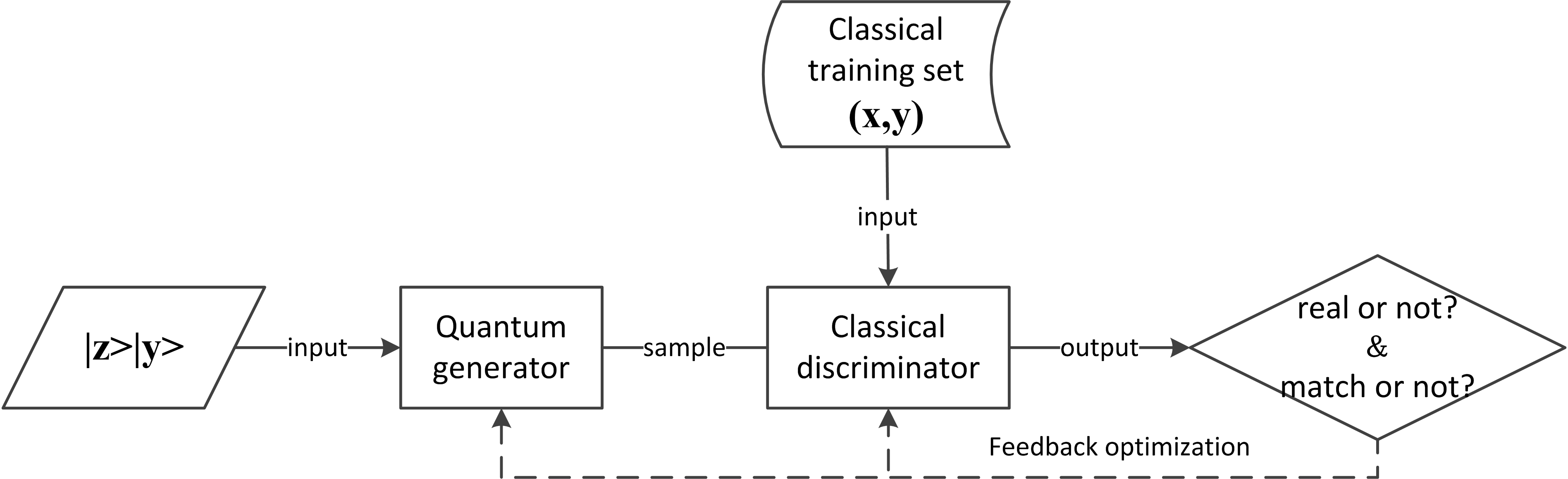}
  \caption{\csentence{Schematic diagram of quantum conditional generative adversarial network.}}
  \end{figure}

\subsection{Entangled state coding of conditional information and circuit design}
For the quantum scheme of CGAN, an important topic is how to input the classical conditional variables into the quantum generator, which involves the quantum state encoding of the conditional variables and the circuit design for preparing this quantum state. In this paper, taking the representative category labels in the conditional variables as an example, the method of coding the entangled state of conditional information and designing corresponding circuit are explained in detail.

As shown in Fig. 4, the real data input to the discriminator are the data pairs $\left( {x,y} \right)$ sampled from the classical training set, where $y$ represents the conditional variable. The generator obtains the representation method of the conditional variables and the probability distribution of various samples in the training set through $\left| {\rm{y}} \right\rangle $. Therefore, $\left| {\rm{y}} \right\rangle $ is a quantum state entangled by $m$-categories conditional variables according to the probability distribution of real samples
\begin{eqnarray}\label{eqexpmuts}
\left| y \right\rangle  = \sum\limits_{j = 1}^m {\frac{1}{{{\alpha _j}}}\left| {{y_j}} \right\rangle } ,
\end{eqnarray}
where $1/{\alpha _j} = {\left( {p\left( {x\left| {{y_j}} \right.} \right)} \right)^{{\rm{ - }}1/2}}$, and ${1/{\alpha _j}}$ meets the normalization conditions: ${\sum\limits_{j = 1}^n {\left| {1/{\alpha _j}} \right|} ^2} = 1$.

The category labels of classical data samples used for machine learning tasks are generally coded by one-hot method. Assuming that three categories of data are generated, and the classical binary representations of three labels are: $001, 010, 100$. Since the classical discriminator will perform classification processing on the generative distribution and the real distribution, it is most reasonable to use the same one-hot method to encode $\left| {{{\rm{y}}_j}} \right\rangle $. It also happens to be similar in form to the quantum three-particle $W$ state, ${\left| W \right\rangle _3} = 1/3\left( {\left| {001} \right\rangle  + \left| {010} \right\rangle  + \left| {100} \right\rangle } \right)$. When designing a quantum circuit to prepare $\left| {\rm{y}} \right\rangle $, the quantum circuit of preparing a multi-particle $W$ state can be used as a template, which reduces the complexity of circuit design to a certain extent.

Taking $\left| y \right\rangle  = {\left| W \right\rangle _3}$ as an example, where $m=3$, ${\alpha _j} = \sqrt 3 \left( {j = 1,2,3} \right)$, which means that the training set contains three categories of uniformly distributed data. The specific preparation process of ${\left| W \right\rangle _3}$ can be divided into two steps, and the corresponding quantum circuit is shown in Fig. 5. The first step is to use a combination of single qubit rotation gates and CNOT gate. By adjusting the rotation angle, the qubits are prepared into a special state containing only three terms, i.e.,
\begin{eqnarray}\label{eqexpmuts}
\left| {{Q_b}{Q_c}} \right\rangle :\left| {00} \right\rangle  \to \frac{1}{{\sqrt 3 }}\left( {\left| {00} \right\rangle  + \left| {01} \right\rangle  + \left| {10} \right\rangle } \right).
\end{eqnarray}
According to the calculation rule of quantum circuit cascade, there is a equation
\begin{eqnarray}\label{eqexpmuts}
EDCBA{\left[ {1,0,0,0} \right]^{\rm T}} = \frac{1}{{\sqrt 3 }}{\left[ {1,1,1,0} \right]^{{\rm T}\;}}.
\end{eqnarray}
By solving this equation, the parameters ${\theta _1} = {\theta _3} = {\rm{0}}{\rm{.55357436}}, {\theta _2} = {\rm{ - 0}}{\rm{.36486383}}$ in the quantum circuit can be obtained. The second step is to select the quantum gates without parameters to design circuit. Firstly perform the NOT gate (i.e., Pauli X gate) on $\left| {{Q_b}} \right\rangle $ and $\left| {{Q_c}} \right\rangle $, then apply the Toffoli gate to set the $\left| {{Q_a}} \right\rangle $ equal to $\left| 1 \right\rangle $, when $\left| {{Q_b}} \right\rangle $ and $\left| {{Q_c}} \right\rangle $ equal to $\left| 1 \right\rangle $. Finally, perform a NOT gate on $\left| {{Q_b}} \right\rangle $ and $\left| {{Q_c}} \right\rangle $ to restore the state at the end of the first step. After the above operations, the initial state $\left| {000} \right\rangle $ can be evolved into ${\left| W \right\rangle _3}$.

  \begin{figure}[h!]
  \includegraphics[width=4.5in]{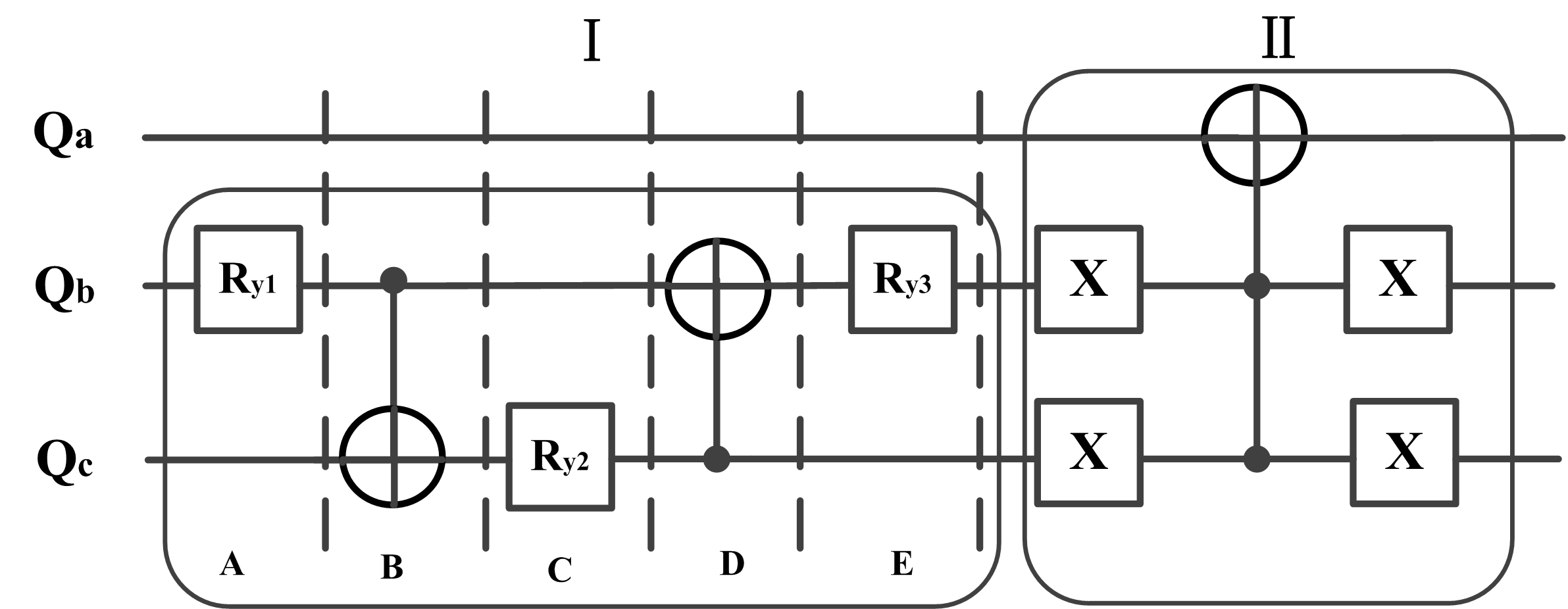}
  \caption{\csentence{The quantum circuit for preparation of three-particle W-state quantum circuit.}}
      \end{figure}

Using the one-hot method to encode the conditional information in the quantum state requires relatively more quantum resources, but it can reduce the workload of converting the data into other encoding forms when the data is classically post-processed. When designing the circuit for preparing quantum state of the conditional information, as long as the fixed template is followed, the parameter value is obtained by changing the probability amplitude on the right end of Eq. 5, and the multi-class label information that meets any probability distribution can be expressed.

\subsection{Circuit design of quantum generator}
Quantum computing forms a quantum circuit through the arrangement and combination of wires and basic quantum gates, which act on the quantum state to achieve the evolution of the system. The so-called parameterized quantum circuit is to choose a combination of parameterized quantum rotation gates and other quantum logic gates to constitute the circuit. Single-qubit gates are used to realize qubit rotation, while multi-qubit gates mainly realize entanglement between qubits. Representing the quantum state and the quantum gate in the form of a vector and a unitary matrix, it means that the mathematical connotation of the quantum gate operation is linear transformation, which is similar to classical machine learning. In that, the role of parameters in PQCs and classical neural networks is consistent.

Due to the unitary constraints of quantum gates, to generate $N$ bits of data, $N = {N_d} + {N_c}$ qubits resources are required, where ${N_d}$ channels process sample data and ${N_c}$ channels receive conditional information. For the quantum generator, the input ${\left| 0 \right\rangle ^{ \otimes {N_d}}}\left| y \right\rangle $ is converted into the final state ${\left| x \right\rangle _G}\left| y \right\rangle $ after the ${L_G}$ layers combination unitary operations, where the ${\left| x \right\rangle _G}$ represents the generative distribution. Sampling the final state of the generator can collapse the quantum state to classical data. The quantum generator is realized by a PQC based on quantum gate computing mechanism, which is composed of rotation layers and entanglement layers alternately arranged. Due to the unitary nature of the quantum gate set, if the rotation layer and the entanglement layer alternately perform operations and form a sufficiently long layer sequence, any unitary transformation can be performed on the initial state in theory.

According to the decomposition theorem of single qubit unitary operation, a single rotation layer is composed of two $R_z$ gates and one $R_x$ gate arranged at intervals, that is $\prod\limits_{i = 1}^N {R_z^{}\left( {\theta _{l,3}^i} \right)R_x^{}\left( {\theta _{l,2}^i} \right)R_z^{}\left( {\theta _{l,1}^i} \right)} $. The superscript $i$ indicates that the quantum gate acts on the $i$-th qubit, and the subscript $l$ indicates that the operations perform on the $l$-th layer. The matrix representation of $R_x$ gate and $R_z$ gate are
\[{R_x}\left( \theta  \right) = \left[ {\begin{array}{*{20}{c}}
{\cos \left( {\theta /2} \right)}&{ - i\sin \left( {\theta /2} \right)}\\
{ - i\sin \left( {\theta /2} \right)}&{\cos \left( {\theta /2} \right)}
\end{array}} \right],{R_z}\left( \theta  \right) = \left[ {\begin{array}{*{20}{c}}
{{e^{ - i\theta /2}}}&0\\
0&{{e^{i\theta /2}}}
\end{array}} \right].\]

A single entanglement layer generally selects two-qubit controlled rotation gates (such as CRX, CRY, CRZ gate) and general two-qubits logic gates (such as CNOT gate) for permutation and combination. The arrangement of quantum gates is related to the connectivity among qubits, thus affecting the expressiveness and entanglement capabilities of PQCs. There are three common connection topologies among qubits: circle, star, and all-to-all connectivity \cite{Sim2019,Linke}. For circle or star connectivity, the entanglement between certain qubits will not occur in a single layer, which means that more layers are required to fit the distribution of complex targets. This phenomenon undoubtedly increases the difficulty of parameters optimization. All-to-all connectivity is an ideal topology structure among qubits. Although the number of parameters of a single-layer will exceed the other two methods, a shallow all-to-all connectivity quantum circuit can achieve better generative results and the computational overhead of algorithm is cheaper.

When designing the PQC of quantum generator, it is necessary to ensure that the qubits are fully connected. According to the above rules, the quantum generator circuit of QCGAN is shown in Fig. 6. The "XX" in the Fig. 6 represents an operation involving two qubits, where any one is the control qubit, and the other is the target qubit. When the control qubit is $\left| 1 \right\rangle $ or $\left| 0 \right\rangle $ (specified by the operation), the target qubit is operated accordingly. The ${N_c}$ qubits are only responsible for transmitting conditional information to the other ${N_b}$ qubits, and continue to pass the conditional information to the discriminator in post-processing. Therefore, no rotation operation is performed on them, and they are only used as control qubits to affect the circuit for data generation.

  \begin{figure}[h!]
  \includegraphics[width=4.5in]{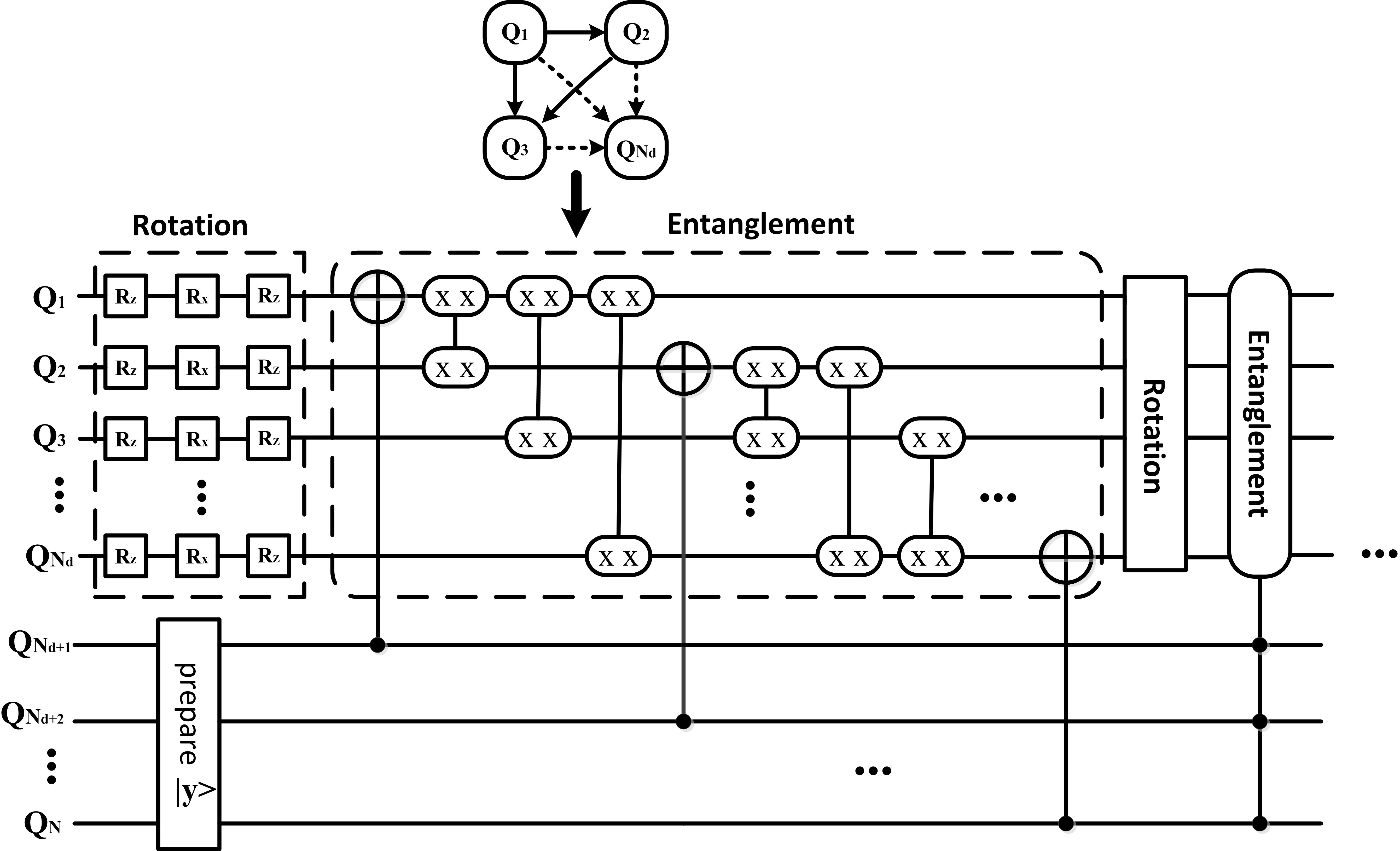}
  \caption{\csentence{The template of quantum generator circuit.}}
      \end{figure}

\subsection{Adversarial training strategy}
The training of the QCGAN is a parameter optimization quantum algorithm with a feedback loop. The parameters of quantum generator and classical discriminator are denoted by $\theta $ and $\phi $, respectively. Similar to the classical CGAN, the objective function of QCGAN is
\begin{eqnarray}\label{eqexpmuts}
\mathop {\min }\limits_{{G_\theta }} \mathop {\max }\limits_{{D_\phi }} V\left( {D,G} \right) = {E_{x \sim {p_{real}}}}\left[ {\log D\left( {x\left| y \right.} \right)} \right] + {E_{x \sim {p_\theta }}}\left[ {\log \left( {1 - D\left( {{x_G}\left| y \right.} \right)} \right)} \right].
\end{eqnarray}

At the beginning of training, all parameters in quantum circuit and binary classification neural network are given random initial values. During the adversarial training process, the parameters of generator and discriminator are alternately optimized. The parameters of the quantum generator circuit are fixed first to optimize the parameters of the discriminator. The discriminator simultaneously judges the randomly sampled batch training data and the data sampled from the quantum generator. The output value of the discriminator represents the probability that the corresponding input comes from the real distribution, and the gradient is calculated in the direction of maximizing the objective function of discriminator to optimize the parameter $\phi $. Modifying the parameters of discriminator and repeating the above optimization operations, so that discriminator can not only learn the characteristics of real data distribution but also have the ability to discriminate the data from generative distribution. Then the parameters of discriminator are fixed, and the input of discriminator is only the results of the generator sampling. The larger the output of the discriminator, the smaller the gap between the generative distribution and the previously learned real distribution. In that, the gradient is calculated according to the direction of maximizing the objective function of generator to optimize the parameters $\theta $. The ability of generator to fit the true distribution is continuously improved by modifying the parameters and repeating the circuit on the quantum computing device. The alternate optimization of generator and discriminator parameters must be iteratively performed until generator can reconstruct the state distribution of the training set.

According to the above connotation of adversarial training, Eq. 6 is decomposed into the unsaturated maximization objective function that generator and discriminator obeys respectively,
\begin{eqnarray}\label{eqexpmuts}
\left\{ {\begin{array}{*{20}{l}}
{\max {V_{{D_\phi }}} = {E_{x \sim {p_{real}}}}\left[ {\log D\left( {x\left| y \right.} \right)} \right] + {E_{x \sim {p_\theta }}}\left[ {\log (1 - D\left( {{x_G}\left| y \right.} \right))} \right]}\\
{\max {V_{{G_\theta }}} = {E_{x \sim {p_\theta }}}\left[ {\log \left( {D\left( {{x_G}\left| y \right.} \right)} \right)} \right]}
\end{array}.} \right.
\end{eqnarray}
During the training process, the gradient descent method is used to optimize the parameters. This method needs to calculate the gradient information ${\nabla _\theta }{V_{{G_\theta }}}$ and ${\nabla _\phi }{V_{{D_\phi }}}$. For classical neural networks, backpropagation can be used directly to calculate the gradient value of the objective function effectively. But for quantum devices, only the measurement results can be obtained, in that the output probability of discriminator cannot be directly accessed. Therefore, the gradient estimation of a parameterized quantum circuit needs to follow the theorem: for a circuit containing the parameter unitary gates $U\left( \eta  \right) = {e^{ - \frac{i}{2}\eta \Sigma }}$, the gradient of the expectation value of an observable $B$ with respect to the parameter $\eta $ reads
\begin{eqnarray}\label{eqexpmuts}
\frac{{\partial {{\left\langle B \right\rangle }_\eta }}}{{\partial \eta }} = \frac{1}{2}\left( {{{\left\langle B \right\rangle }_{{\eta ^ + }}} - {{\left\langle B \right\rangle }_{{\eta ^ - }}}} \right).
\end{eqnarray}
The ${\left\langle  \right\rangle _{{\eta ^ \pm }}}$ in Eq. 8 represents expectation value of observable with respect to the output quantum wave function generated by the same circuit with parameter ${\eta ^ \pm } = \eta  \pm \frac{2}{\pi }$ \cite{LiuJ}. This is an unbiased estimation method for the gradient of PQC. According to this theorem, the gradient of the output of the discriminator with respect to the parameters $\theta $ can be calculated
\begin{eqnarray}\label{eqexpmuts}
\frac{{\partial {V_{{G_\theta }}}}}{{\partial {\theta _i}}} = \frac{1}{2}{E_{x \sim {p_{{\theta ^ + }}}}}\left[ {\log D\left( {x\left| y \right.} \right)} \right] - \frac{1}{2}{E_{x \sim {p_{{\theta ^ - }}}}}\left[ {\log D\left( {x\left| y \right.} \right)} \right],
\end{eqnarray}
where ${\theta ^ \pm } = \theta  \pm \frac{2}{\pi }{e^i}$ and ${e^i}$ represents the $i$-th unit vector in the parameter state space, i.e., ${\theta _i} \leftarrow {\theta _i} \pm \frac{2}{\pi }$. In order to estimate the gradient of each parameter, every single parameter needs to be optimized and then evaluated repeatedly. In the case of small-scale numerical simulation, the wave function can be used to directly calculate the expectation value. Another method is to calculate the probability distribution based on the wave function, and then sample the gradient for estimation \cite{Situ}.

\section{Performance evaluation}
In order to evaluate the performance of the algorithm proposed in this paper, the classical GAN \cite{Good} and CGAN \cite{Mirza}, QGAN \cite{Seth} and QCGAN are mainly compared from the perspectives of time complexity and algorithm function. The performance comparison of the four generative adversarial algorithms is shown in Table 1.

In the classical CGAN algorithm, the process of generator parameters optimization can be seen as performing gradient descent in the convex set of the normalized covariance matrix of the data set to fit the real distribution. Therefore, the time complexity of generating data that fit the $N$-dimensional classical distribution is $O({N^2})$. In contrast, the time complexity of a quantum information processor to perform a linear transformation on an $N$-dimensional vector is $O(N)$. Even if optimizing the each parameter needs to modify and execute the PQC twice, the calculation time complexity of QCGAN is still lower than that of CGAN when the same parameter optimization strategy is adopted (neglecting the time cost of preparing classical data into quantum states). On the other hand, the classical CGAN algorithm cannot directly generate discrete data due to the influence of differentiable constraints during parameter optimization, while QGAN can directly generate discrete data and also has the ability to generate continuous distribution \cite{1901.00848}. In addition, the QCGAN algorithm proposed in this paper directly encodes classical data in quantum state, so its resource consumption is ${N_d} + {N_c}$ the same as classical CGAN (where ${N_d}$ is the resource consumption of generating target data, and ${N_c}$ is the conditional information resource consumption). While the resource consumption of unsupervised GAN and QGAN algorithms is $N$, which is equal to the generative target data size.

Compared with unconstrained QGAN, the input of conditional information brings prior knowledge about the training set to the model, turning unsupervised QGAN into a weakly supervised or supervised adversarial learning model, thereby achieving controllable data generation process. The learning results of unconstrained QGAN are more inclined to present the average state of all data in training set. But due to adding the conditional information, QCGAN will accordingly show an advantage in the fitness of the generated results to the real distribution. Moreover, the generator trained by QGAN is still purposelessly generated, which can only guarantee the authenticity of the generated data but cannot expand other functions. While QCGAN can achieve different purpose generation tasks by introducing different conditional information, which can fully reflect the subjective initiative of people and realize the interaction between people and algorithms. It can be considered that QCGAN is a human-centered algorithm. Therefore, from a functional perspective, the generators trained by QCGAN have more extensive application scenarios and higher efficiency.

\begin{table}[h!]
\caption{Performance comparison of $4$ generative adversarial network algorithms}
\begin{tabular}{ccccc}
\hline
Algorithm name                 & GAN        & CGAN             & QGAN                   & QCGAN                  \\ \hline
Time complexity                & $O({N^2})$ & $O({N^2})$       & $O({N})$               & $O({N})$               \\
Generator resource consumption & $N$ bits   & $N_d + N_c$ bits & $N$ qubits             & $N_d + N_c$ qubits     \\
Generate data type             & Continuous & Continuous       & Continuous \& Discrete & Continuous \& Discrete \\
Whether human-center algorithm & No         & Yes              & No                     & Yes                      \\ \hline
\end{tabular}
\end{table}

\section{Experimental}
In this paper, the synthetic BAS$(2, 2)$ (Bars and Stripes) data set is used for the experiments and analyses of the classical data classification generation task. The TensorFlow Quantum (TFQ), an open source quantum cloud computing platform for the rapid prototyping of hybrid quantum-classical models for classical or quantum data \cite{TFQ}, is introduced to realize the simulation experiments.

\subsection{BAS data set}
The BAS$(m,n)$ data is a composite image containing only horizontal bars or vertical stripes on a two-dimensional grid. For $m \times n$-pixel images, there are only ${2^m} + {2^n} - 2$ valid BAS images in all ${2^{m \times n}}$ cases. This defines the target probability distribution, where the probabilities for valid images are specified constants, and the probabilities for invalid images are zero. The generation goal of the experiment is the classical data of BAS$(2, 2)$, which seem to be a insufficient challenging for quantum computers intuitively. However, the effective quantum state represented by the BAS$(2,2)$ data set have a minimum entanglement entropy of ${S_{BAS\left( {2,2} \right)}} = {\rm{1}}{\rm{.25163}}$ and a maximum achievable entropy of ${S_{BAS\left( {2,2} \right)}} = {\rm{1}}.{\rm{79248}}$, which is the known maximum entanglement entropy of four-qubit states set \cite{Higu}. Therefore, the data have rich entanglement properties and are very suitable as a generation target for quantum adversarial training.

The BAS$(2,2)$ images in the training set are divided into three categories. The horizontal bar images and the vertical stripe images are respectively one category, and the image with pixel values of all $0$ or all $1$ is the other category. And the effective BAS images conform to the uniform distribution. According to the classification standard, the category labels are one-hot encoded and added to the basic data set as the conditional information. Hence the generator require $7$ qubits resources, as processing the pixel information of BAS data requires $4$ qubits, receiving conditional information requires $3$ qubits.

\subsection{Experimental setup}
The codes synthesis $6000$ samples to form the training set, including three categories of BAS data (a total of $6$ valid images) that meet the above requirements and their category labels. During training, all data is out of order firstly, and then extracted by batch size. For the pre-training of the BAS data set, the discriminator and generator are alternately trained once in each iteration optimization. The batch size of each training is $40$, and there are totally $100$ epochs for iterative training. In each epoch, iterative training the network 150 times, so that the discriminator can traverse the entire training set. Considering that the improper setting of the learning rate will cause the network gradient to disappear/explode, setting the learning rate $\times 0.1$ to reduce it every $10$ epochs of training. The Adam (Adaptive Moment
Estimation) optimizer provided by the open source library is introduced for both generator and discriminator, and the initial learning rate is set as $0.001$.

Each epoch of training optimization completes, the output of generator is sampled to inspect the quality of the current generation distribution. The inspection mainly including three points:

$(1)$ whether the generated pixel data constitutes a valid BAS image;

$(2)$ whether the generated pixel data matches the conditional information;

$(3)$ whether the generated all data conforms to the uniform distribution.

Since the training process of the adversarial network is relatively unstable, if the comprehensive accuracy of the above three investigation points reaches the preset threshold of $95\% $, the training process can be chosen to terminate early. If the threshold can not be reached all the training time, $100$ epochs of alternate training are performed according to the preset settings, and then analyze the convergence of the objective function in the whole training process. After that, the adversarial network can be trained again after reasonable adjustments to the training strategy and hyperparameters, by summarizing the reasons for the unsatisfactory training results.

\section{Results and discussion}
In the simulation process, a series of comparative experiments are conducted on the performance of the generator using circle, star, and all-to-all connected quantum circuits firstly. The results verified the superiority of designing an all-to-all connected topology of the quantum generator in this scheme. According to the result of the comparative experiment, the PQC structure shown in Fig. 7 is used as the generator of QCGAN. The input $\left| y \right\rangle $ of the generator is ${\left| W \right\rangle _3}$, which is prepared in advance with the circuit shown in Fig. 5.

The discriminator is classical so it is implemented using the classical deep learning framework, TensorFlow, which can form a hybrid quantum-classical model with TFQ. The discriminator has one input layer with dimension ${N_{\rm{d}}} + {N_c} = 7$, one hidden layer made up of $4$ neurons and one output neuron. Since the discriminator directly judges the expectation value of the generator output, the hidden layer selects the linear ReLU activation function.

  \begin{figure}[h!]
  \includegraphics[width=4.5in]{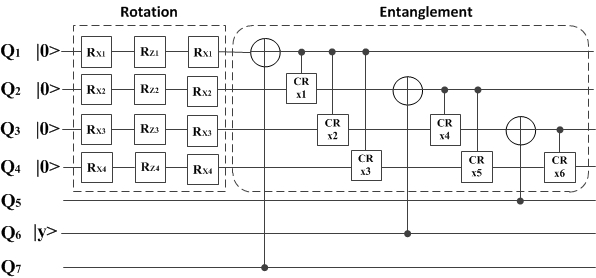}
  \caption{\csentence{The quantum  generator circuit diagram in this QCGAN experiment.}}
  \end{figure}

As shown in Fig. 8, in the overall trend, the loss function value of the discriminator gradually decreases and the loss function value of the generator gradually increases. After training, both the losses of generator and discriminator converge to near the expected equilibrium point. As the epoch of training increases, the model gradually stabilizes and the relationship between generator and discriminator is more intense. So it shows in Fig. 8 that there is still a large oscillation around the expectation value after the convergence. This phenomenon is also related to the influence of noise on quantum systems which access through cloud platform.

  \begin{figure}[h!]
  \includegraphics[width=4.5in]{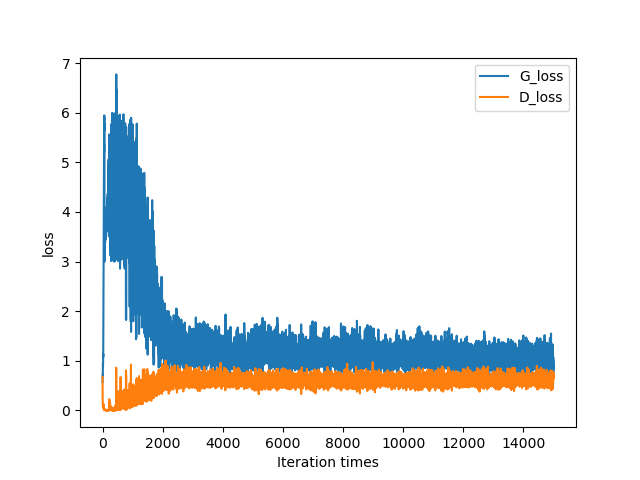}
  \caption{\csentence{The discriminator (in orange) and generator (in blue) loss with respect to iterations.}}
      \end{figure}

After the pre-training of the BAS data set is completed, quantum generator result is sampled $10,000$ times to analyze the generative distribution. The probability distribution of the generated data is shown in Fig. 9(a). It can be seen that most of the generated data fall in the six valid BAS mode images, and the three categories BAS images basically conform to the uniform distribution with $97.15\%$ accuracy. Fig. 9(b) visualizes the first $100$ generative samples in the form of pixel maps of $1$, $70$ and $100$ epoch, which shows that the quantum generator gradually has the ability to generate BAS$(2,2)$ images after pre-training.

  \begin{figure}[h!]
  \includegraphics[width=4.5in]{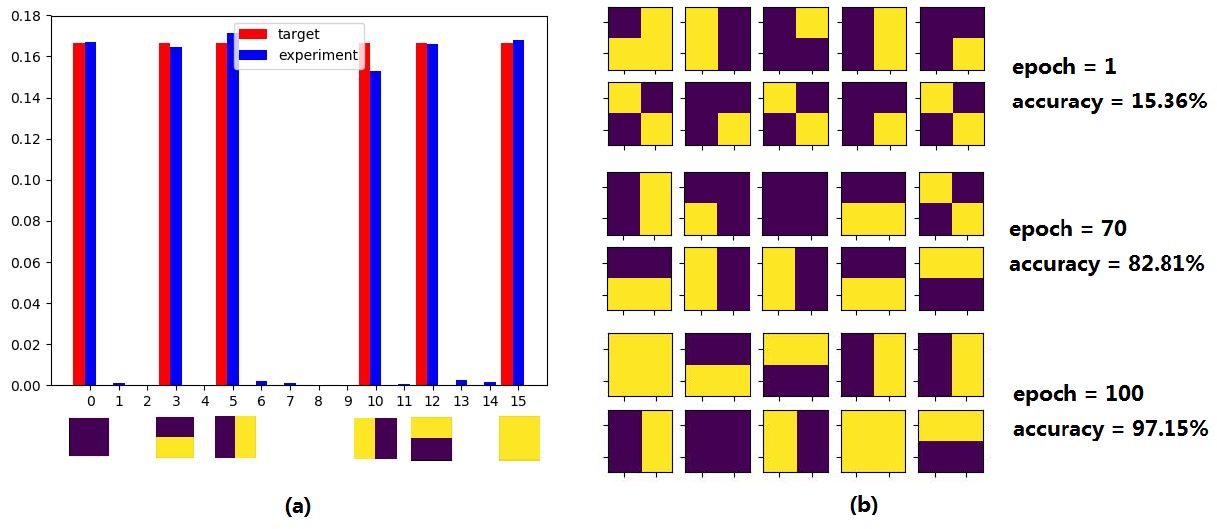}
  \caption{\csentence{$2 \times 2$ Bars-and-Stripes samples generated from QCGAN.}(a)The final probability distribution of the generative BAS data.(b)BAS samples generated from QCGAN with different epoch(For illustrative purpose, we only show 10 samples for each situation.).}
      \end{figure}

The parameters of quantum gates in the optimal generator are extracted after pre-training, and then use the generator circuit shown in Fig. 7 to realize the task of generating classification images. The parameters of PQC in Fig. 5 are adjusted to set the input $\left| y \right\rangle $ as $\left| {001} \right\rangle $, and then sample the output ${\left| x \right\rangle _G}$ of generator. The result shows that two kinds of horizontal stripe images meet the uniform distribution, which means that the quantum generator can generate data of multiple categories that meet the conditional constraints through the guidance of conditional information.

\section{Conclusion}
Combining the classical CGAN algorithm with quantum computing ideas, this paper proposes a quantum conditional generative adversarial network algorithm for human-centered paradigm, which is a general scheme suitable for fitting classical data distribution. This paper gives a detailed interpretation of our design focus, including the configuration design of PQC as the generator, the parameter gradient estimation method of adversarial training strategy as well as the specific steps of the algorithm's cloud computing implementation.

The effect of the QCGAN algorithm is that by adding conditional constraints related to the training data set in the input layer, which effectively guides the network to generate data that meets specific requirements. This step increases the controllability of the generation process, but also more in line with the current human-centered requirements for machine learning algorithms. Compared with classical CGAN, the time complexity of the QCGAN algorithm proposed in this paper is lower, and it is more in line with the needs of actual application scenarios. Through experiments on the quantum cloud computing platform, the results show the QCGAN can generate the BAS data distribution effectively and the generator of QCGAN can output correct data guided by the conditional constraint in cloud.

Given that QGAN has the ability to generate discrete data and the potential to dig out data distributions that cannot be effectively summarized by classical calculations, QGAN and classical GAN are functionally complementary. Many known variants of GAN can generate very realistic images, audio, and video, in that the combination of these algorithms and quantum mechanics is undoubtedly the icing on the cake. Our future work will focus on the quantum schemes of some classical GAN variant algorithms and constructing quantum machine learning algorithms that conform to the HCC paradigm and the corresponding cloud computing implementation.


\begin{backmatter}
\section*{Abbreviations}
QGAN: Quantum generative adversarial network; QCGAN: quantum conditional generative adversarial network; NISQ: Noisy Intermediate-Scale Quantum; CGAN: Conditional generative adversarial network; HCC: human-centered computing; GAN: Generative adversarial network; PQC: Parameterized quantum circuit; TFQ: TensorFlow Quantum; BAS: Bars and stripes

\section*{Acknowledgements}
This work is supported by National Natural Science Foundation of China (Grant Nos. 62071240 and 61802002); the Natural Science Foundation of Jiangsu Province (Grant No. BK20171458); the Graduate Research and Practice Innovation Program of Jiangsu Province (Grant No. KYCX20\_0969); the Natural Science Foundation of Jiangsu Higher Education Institutions of China under Grant No.19KJB520028; the Priority Academic Program Development of Jiangsu Higher Education Institutions (PAPD).

\end{backmatter}
\end{document}